\documentclass[conference]{IEEEtran}
\makeatletter
\newcommand{\newlineauthors}{%
  \end{@IEEEauthorhalign}\hfill\mbox{}\par
  \mbox{}\hfill\begin{@IEEEauthorhalign}
}
\makeatother
\IEEEoverridecommandlockouts
\usepackage{algorithmicx}
\usepackage{algorithm}
\usepackage{algpseudocode}
\usepackage{tabularx}
\usepackage{multirow}
\usepackage{xcolor}
\usepackage{float}
\usepackage{booktabs}
\usepackage[caption=false]{subfig}
\usepackage{cite}
\usepackage{amsmath,amssymb,amsfonts}
\usepackage{cleveref}

\usepackage{graphicx}
\usepackage{textcomp}
\usepackage{xcolor}
\def\BibTeX{{\rm B\kern-.05em{\sc i\kern-.025em b}\kern-.08em
    T\kern-.1667em\lower.7ex\hbox{E}\kern-.125emX}}
\begin{document}

\title{Contour Refinement using Discrete Diffusion in Low Data Regime\\
}

\author{
\IEEEauthorblockN{Fei Yu Guan}
\IEEEauthorblockA{
\textit{Dept. of Mathematical and}\\\textit{ Computational Sciences}\\
\textit{University of Toronto}\\
Toronto, Canada\\
vi.guan@mail.utoronto.ca
}

\and \IEEEauthorblockN{Ian Keefe}
\IEEEauthorblockA{
\textit{Institute of Aerospace Studies}\\
\textit{University of Toronto}\\
Toronto, Canada\\
ian.keefe@robotics.utias.utoronto.ca
}

\and \IEEEauthorblockN{Sophie Wilkinson}
\IEEEauthorblockA{
\textit{School of Resource \&}\\ \textit{Environmental Management}\\
\textit{Simon Fraser University}\\
Burnaby, Canada\\
sophie\_wilkinson@sfu.ca
}
\newlineauthors

\and \IEEEauthorblockN{Daniel D.B. Perrakis}
\IEEEauthorblockA{
\textit{Pacific Forestry Centre}\\
\textit{Natural Resources Canada}\\
Victoria, Canada\\
daniel.perrakis@nrcan-rncan.gc.ca
}
\and \IEEEauthorblockN{Steven L. Waslander}
\IEEEauthorblockA{
\textit{Institute of Aerospace Studies}\\
\textit{University of Toronto}\\
Toronto, Canada\\
steven.waslander@robotics.utias.utoronto.ca\\
}}

\maketitle

\begin{abstract}
Boundary detection of irregular and translucent objects is an important problem with applications in medical imaging, environmental monitoring and manufacturing, where many of these applications are plagued with scarce labeled data and low in situ computational resources. While recent image segmentation studies focus on segmentation mask alignment with ground-truth, the task of boundary detection remains understudied, especially in the low data regime. 

In this work, we present a lightweight discrete diffusion contour refinement pipeline for robust boundary detection in the low data regime. We use a Convolutional Neural Network(CNN) architecture with self-attention layers as the core of our pipeline, and condition on a segmentation mask, iteratively denoising a sparse contour representation. We introduce multiple novel adaptations for improved low-data efficacy and inference efficiency, including using a simplified diffusion process, a customized model architecture, and minimal post processing to produce a dense, isolated contour given a dataset of size $<$500 training images. Our method outperforms several SOTA baselines on the medical imaging dataset KVASIR, is competitive on HAM10K and our custom wildfire dataset, Smoke, while improving inference framerate by 3.5X. 
 \end{abstract}

\section{Introduction}
\label{sec:intro}

Boundary detection is an important subproblem of image segmentation that focuses on precisely identifying the pixel-level boundary surrounding a region of interest.  It is very important for numerous tasks, such as tumor identification in medical imaging, defect detection in manufacturing and wildfire front detection. Despite these important applications, boundary detection is an understudied computer vision problem relative to more mainstream tasks such as segmentation and object detection. Boundary detection usually involves a small number of labeled images for data privacy reasons, leading to challenges in generalization due to the small dataset size. 

Existing methods for boundary detection fall into four categories: non-learning-based segmentation approaches, CNN methods, foundation models and generative methods. Non-learning-based methods such as Edgeflow~\cite{b4} compute edge flow vectors to detect intensity or texture changes, but have not been tested in handling translucent object boundaries or noisy backgrounds. Geodesics Active Contours~\cite{b5} follows the boundary of an object by minimizing the "energy" (the area of flat regions in the object). CNN approaches significantly improve segmentation mask accuracy over non-learning based techniques. Works such as BDCN~\cite{b37}, HED~\cite{b38} produce fine contours directly from images using CNNs. In parallel, segmentation networks tailored to specific domains, such as Deep Smoke Segmentation ~\cite{b9} and real-time fire segmentation ~\cite{b10}, demonstrate the benefits of end-to-end learning for challenging visual phenomena. %CNN models such as EfficientNet~\cite{tan2020efficientnetrethinkingmodelscaling} excel in popular classification datasets such as ImageNet~\cite{5206848} and CIFAR-10~\cite{krizhevsky2009learning} (see~\cite{Dumitru_2023}), suggesting that there is still an inherent potential for efficient CNNs, despite the introduction of foundation models. 
DUCKNet~\cite{b12} is another such example where CNNs outperform other methods in polyp segmentation ~\cite{b33}.

\begin{figure}[t]
\centering
\includegraphics[width=0.8\linewidth]{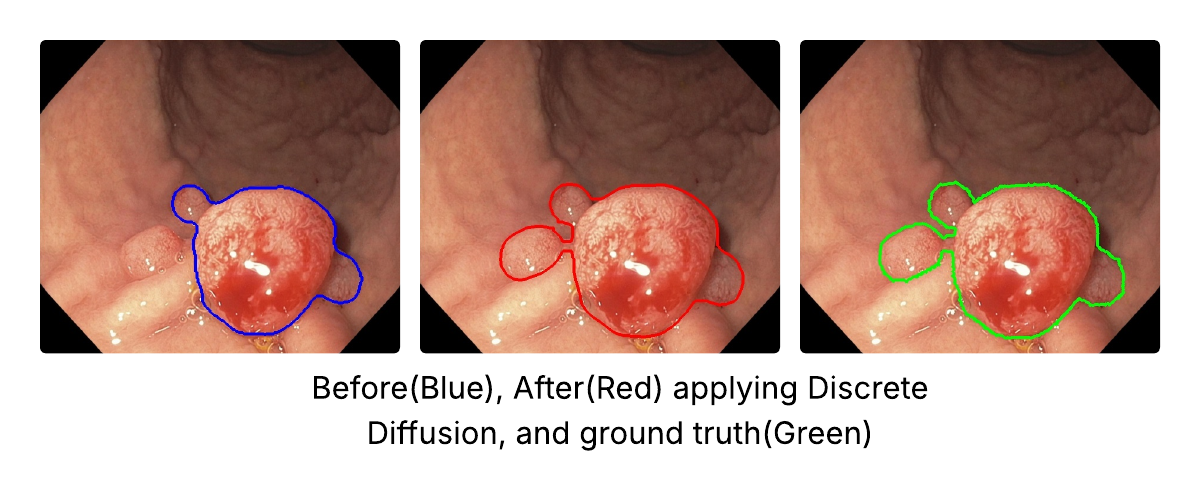}
\caption{Our contour refinement pipeline on the KVASIR dataset.}
\label{fig_pipeline}
\end{figure}

The recent introduction of segmentation foundation models, notably Segment Anything 2 (SAM2) ~\cite{b1}, has shifted the paradigm toward prompt-based object segmentation. These models can generate masks from an image and minimal user inputs (e.g., points or boxes) and can generalize across domains without finetuning. However, their accuracy is highly sensitive to prompt quality, and their performance degrades for translucent objects such as smoke, fire, and certain medical boundaries. Moreover, in low-data regimes where training or fine-tuning is constrained, prompt generation itself becomes a bottleneck, as manual annotation is laborious and automatic detectors are prone to failure.

Generative models have also been applied to segmentation and boundary detection. Some generative methods ~\cite{b41} use  Generative Adversarial Networks (GANs)~\cite{b17} for segmentation.  However, GANs~\cite{b17} are prone to training sensitivity and mode collapse. Diffusion models have recently emerged as powerful generative or refining tools, producing high-quality outputs through iterative denoising. Generative methods~\cite{b6, b40} have employed diffusion models for segmentation mask generation. Moreover,  {SegRefiner~\cite{b2} has demonstrated that discrete diffusion models can refine segmentation masks through iterative denoising on large datasets, but adaptations for low-data settings and translucent object contours remain unexplored. %Works like ContourDiff~\cite{chen2024contourdiffunpairedimagetoimagetranslation} improve the image translation by taking contours as inputs. 

In this work, we introduce a discrete diffusion process built upon an attention-based DUCKNet ~\cite{b12}, a CNN based backbone with modifications to improve convergence and robustness under data scarcity. This choice of network is signficantly less computationally complex when compared to other SOTA methods for boundary segmentation. We quantize the confidence score into categories, incorporate loss terms specifically for improving low data generalization, and use a morphological function called Skeletonize~\cite{b16} for refining pixel level contours from the network outputs.

Our contributions are as follows:
\begin{itemize}
\item A computationally efficient discrete diffusion contour refinement pipeline for translucent object boundaries in low-data regimes.

\item A collection of low-data training optimizations, including a quantized discrete confidence score for improved quality of the outputs, specific loss function terms to speed up convergence, and a morphological post-processing method to ensure dense, closed contours.

\item Through extensive evaluations on three datasets, HAM10K ~\cite{b18}, KVASIR ~\cite{b19}, and SMOKE, a novel curated smoke detection dataset, we demonstrate notable improvements over several single staged segmentation architectures in both boundary alignment and shape similarity metrics.

\end{itemize}

\section{RELATED WORK}

\subsection{Contour Refinement}
Contour refinement has been proposed as a way to enhance segmentation masks from a coarse initial boundary estimate.  SegFix~\cite{b27} learns to predict fine contours given coarse contours and iteratively refines the boundary at the pixel level.  BPR~\cite{b39}  takes the patches on an object's boundary, refines them, and then reconstructs the boundary, thereby refining both the boundary and the segmentation mask. These methods assume sizeable datasets are available for supervised training and emphasize the segmentation quality as the final output. In contrast, our work focuses on the boundary detection problem specifically in the low data regime, inverting the paradigm used in these works by refining the boundary estimate from an initial segmentation mask. %and applies an adaptation of the diffusion process used in SegRefiner~\cite{wang2023segrefinermodelagnosticsegmentationrefinement} to perform boundary refinement optimized for the low data regime. 

\subsection{Contour Detection}

Contour extraction has also been studied directly in various forms.  Early works have explored using boundaries to obtain  segmentation regions\cite{b30, b28, b4, b32,b7}. More recently, deep learning methods have shown significant improvements in performance and generalization ability over prior works.  Some of these methods directly output contour lines from convolutional networks~\cite{b38, b37}.  To the best of our knowledge, however, the possibility of using generative models to extract and refine contours has not been explored in prior work. 

\subsection{Segmentation of translucent Objects}
Translucence adds an additional challenge to the boundary detection problem, as sharp distinctions between regions do not always exist. Many methods have attempted to solve the segmentation challenge in the presence of translucence.  In Deep Smoke Segmentation~\cite{b9}, a fully convolutional network is trained to segment smoke. In Li ~\cite{b10}, a Deeplab-v3~\cite{b11} variant is used for fire segmentation and produced strong results with real-time inference. DUCKNet~\cite{b12} is a unique architecture for segmenting polyps which we use in this work. DUCKNet uses an additional CNN downsampling path, retaining features of the image at different sizes, adding those features to the main down block to preserve meaningful information that can be lost in a traditional UNET\cite{b38}. Other works \cite{b35, b36} were specifically designed for segmenting polyps / colons. One such approach\cite{b33} designed for the KVASIR benchmark~\cite{b19} uses a modern attention mechanism. All of these existing methods focus on producing a dense mask rather than accurate contours. 
\section{METHODOLOGY}

\subsection{Network Design, Training and Inference Pipelines} 

\begin{figure*}[t]
\centering
  \includegraphics[width=0.8\linewidth]{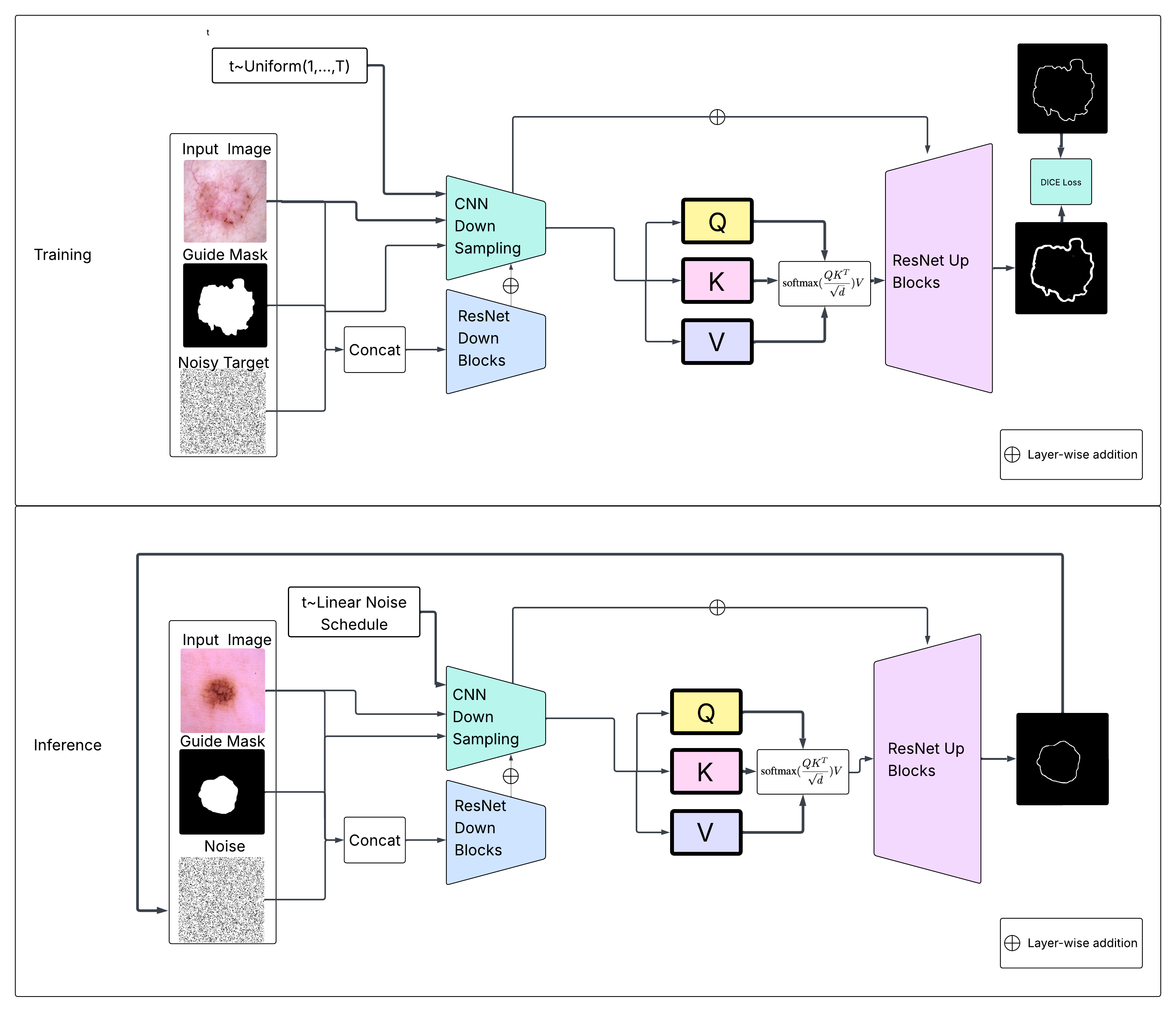}
\caption{An overview of the network architecture, training and inference pipelines.}
\label{fig_pipeline}
\end{figure*}

We use an attention~\cite{b14}-based DUCKNet~\cite{b12} model as the base model for the diffusion. DUCKNet~\cite{b12}  uses an encoder-decoder architecture with a residual downsampling mechanism. This allows the model to capture and process image information at multiple resolutions, preserving spatial details while extracting hierarchical features. For training, we condition the model with a clean ground truth segmentation mask, the image and multinomial noise; for evaluation, we supply the output from a conventional segmentation model finetuned using a larger subset from 389(for smoke dataset) to 500(for other two). The detector (in our case YOLOv11s~\cite{b13}, DeepLab-v3+~\cite{b25}, or SAM2.1\cite{b1})'s outputs, combined with multinomial noise and the input image and use the longest contour line of the segmentation mask as target. We then follow a simplified discrete diffusion process, based on Austin ~\cite{b15}, to train the model. The inference step involves iteratively feeding the output back to denoise, generating an enhanced outline of the contour. Then we post-process the output by applying a Gaussian blur, and then trace the exterior pixel level boundary of this outline using the Skeletonize function~\cite{b16}.

\subsection{Limited-Data Training Modifications} 

We first finetune a lightweight base segmentation model to provide a conditional mask to guide the contour detection. We experiment with three base model architectures, all finetuned: YOLOv11s~\cite{b13}, DeepLab-v3+~\cite{b25} and SAM2.1~\cite{b1}, using finetuned YOLO11~\cite{b13} as prompt generator.  

For faster convergence and minimal noise artifacts in the outputs, we use a discrete diffusion process, treating each pixel as a one-hot vector. Let $q$ be the probability model for the diffusion process. The forward process for continuous diffusion models such as DDIM~\cite{b19} or DDPM~\cite{b15} is characterized by adding a Gaussian noise to an initial state, $x_{0}$ characterized by the following: 
\begin{equation}   
q(\mathbf{x}_{1:T}|\mathbf{x}_0) := \prod_{t=1}^T q(\mathbf{x}_t|\mathbf{x}_{t-1})
\end{equation}
 \begin{equation}   
q(\mathbf{x}_t|\mathbf{x}_{t-1}) := \mathcal{N}(\mathbf{x}_t; \sqrt{1 - \beta_t}\mathbf{x}_{t-1}, \beta_t \mathbf{I})
\end{equation}
where $\beta_{t}\in (0, 1)$ determines the amount of noise added at time $t$.

In the discrete forward process, noise is added to the target by multiplying by a transition matrix, with entries defined in terms of scheduled $\beta_{t}$.
\begin{equation}\mathbf{x_{t+1}} =  \mathbf{x_{t}}\cdot \mathbf{Q}_{t}= \mathbf{x_{t}}\cdot
\begin{bmatrix}
\mathsf{diag}\left(\mathbf{1-\beta_{t}}\right) & \mathbf{\beta_{t}}\\
\mathbf{0} & 1\\
\end{bmatrix}
\label{eq:mat}
\end{equation}
 The entries of the transition matrix can also be characterized as in Austin et al. ~\cite{b15}:
\begin{equation}\mathbf{Q_{t_{i,j}}} = \mathbf{q\left(x_{t}=j|x_{t-1}=i\right)}
\end{equation}
Since contour lines are parameter efficient to generate, with a dataset of size 200 or 400, and a batch size of 15, category size (size of each one-hot vector) of 8, 11 or 32, our model converges quickly and stabilizes within 100 epochs.

Our losses consist of only the simple loss component (\cref{eq:loss1}) , as in ~\cite{b15}, since we find training with the full KL matching loss requires extensive amounts of data and introduces artifacts in the outputs in our low data regime. This would introduce a large error in our evaluation as the static artifacts cannot be easily removed from the background. Also we found the additional texture loss employed in Wang ~\cite{b2} does not provide significant improvement. In the low data setting, the DICE Loss is used instead of the full KL matching loss used in Discrete Diffusion~\cite{b15} to accelerate convergence. We additionally perturb the $\mathbf{x_{0}Q_{1}...Q_{t}}$ with Gumbel-noise and then apply Softmax, as in ~\cite{b40}.
The loss, in its original form is formulated as 
\begin{equation}
\label{eq:loss1}
    \mathbf{L}(\hat{x}, y)=-log(\Tilde{p}_{\theta}(\hat{x}| y))
\end{equation}

 Since we are using the DUCKNet architecture, we use DICE loss, as proposed in Dumitru ~\cite{b12},
\begin{equation}
\label{eq:loss}
\text{$\mathcal{L}_{DICE}(p, g)$} = 1 - \frac{2 \sum_{i} p_i g_i + \epsilon}{\sum_{i} p_i + \sum_{i} g_i + \epsilon}
\end{equation}

The forward training process is characterized by \textbf{Algorithm 1} below.

\begin{algorithm}
\caption{Simplified Discrete Diffusion}
\begin{algorithmic}[1]
\Require Network $f_\theta$, noise schedule $\sigma$ (total noise $\bar{\sigma}$), data distribution $p_{data}$, token transition matrix $Q_{t}$ defined in \cref{eq:mat}, time $\{0,... T\}$, conditional image and mask $\mathbf{c}$. Initialize $\hat\theta = \theta$. Learning rate $\alpha=0.0001$. $\beta_{0}=0.0001, \beta_{T}=0.02$
\For {i in {0,..., num\_train\_steps}}

    \State $\mathbf{x_0}, \mathbf{c} \sim p_0$, $t \sim \mathcal{U}([0, T])$.
    
    \State $\mathbf{x_{t}}\sim GumbelSoftmax( \mathbf{x_{0}}\mathbf{Q}_{1}...\mathbf{Q}_{t})$
    
        \State Compute:$\mathcal{L}_{DICE}(Sigmoid(f_{\theta}(\mathbf{x}_{t}, \mathbf{c}, t)), \mathbf{{x}_{0}})$ \cref{eq:loss}
    
    \State  $\theta = \theta - \alpha \nabla_\theta \widehat{\mathcal{L}}_{DICE}$.
    
    \State  $\hat\theta = EMA(\theta, \hat\theta, 0.98)$.
    
\EndFor
\State \textbf{Return}  $\hat\theta$ 
\end{algorithmic}
\end{algorithm}

We have found that using the standard reverse process(\cref{eq::rev}) described in Austin et al. \cite{b15} degrades the overall performance, as Skeletonize\cite{b16} is very sensitive to artifacts/unsmooth edges. Hence we simply denoise by iteratively feeding the previous output back to the input starting from pure noise, as denoted in \cref{algo:inf}. Please see \cref{tab:diff_proc} for the specific difference.

\begin{equation}
\mathbf{x}_{t-1} \sim Categorical(\mathbf{x}_{t-1} | \frac{x_t Q_t^{\top} \;\odot\; x_0 \overline{Q}_{t-1}}
         {x_0 \overline{Q}_t x_t^{\top}})
\label{eq::rev}
\end{equation}

% Algorithm for Discrete Diffusion Inference
% Algorithm for Discrete Diffusion Inference for Image Generation

% Algorithm for Discrete Diffusion Non-Markovian DDIM Sampling

% Algorithm for Discrete Diffusion Non-Markovian DDIM Sampling
\begin{algorithm}
\label{algo:inf}

\caption{Inference Algorithm}
\begin{algorithmic}[1]

\State \textbf{Input:} Trained model $f_{\theta}$, condition $\mathbf{c}$= Concatenate(image mask), Dimension of state space$\mathcal{D}$ with $|\mathcal{D}| = n$, number of timesteps $T$, subset of timesteps $\tau = \{\tau_1, \tau_2, \ldots, \tau_S\}$ with $\tau_1 = T$, $\tau_S = 1$, $\alpha = 0.01$
\State \textbf{Output:} Generated image sample $\mathbf{x}_0 \sim p(\mathbf{x}, 0)$
\State Initialize $p(\mathbf{x}, T) = \text{Categorical}(\mathbf{x} \mid 1/n)$
\State Sample $\mathbf{x}_T \sim p(\mathbf{x}, T)$ 
\For{$i = 1$ to $S$}
    \State $t = \tau_i$
    \State $z_{t} = $ Concatenate$(\mathbf{c}, \mathbf{x}_t)$ 
    \State $\mathbf{x}_0^{\text{pred}} = f_{\theta}(\mathbf{z}_t, t)$ 
    \State  $\mathbf{x}_t =Softmax(\frac{\mathbf{x}_0^{\text{pred}}}{\alpha})$
\EndFor

\State \textbf{Return}  $\mathbf{x}_{t}.argmax() > threshold$ 
\end{algorithmic}
\end{algorithm}

\section{RESULTS}
\subsection{General Details}
For evaluation, we use ~\cite{b32} to derive contours from the GT masks. F1 scores are set to have a pixel tolerance threshold of 10, due to the large variance / noise apparent in translucent images. We use 8 category classes as a discrete confidence score for KVASIR \cite{b19}, 11 for HAM10K \cite{b18} and 32 for the smoke dataset as opposed to the 2 used in SegRefiner\cite{b2}, which enables the model to have higher expressive capacity. We found that this enables a denser contour sketch with less need for post processing. We found using a batch size of 15 to be effective in most cases. We use the same model size across both datasets, having DUCKNet\cite{b12} kernels of base size 32, ResNet\cite{b22} blocks, a layer repetition factor of 2, and 4 layers of 8-headed attention\cite{b14} in the mid block. We use a small number of training steps (100) for faster convergence while maintaining the generalization in low data settings. For inference, we use 10 steps, since we use a deterministic sampling approach and linear noise scheduling. We use the same core processing image widths, 352 x 352 for all three datasets, scaling images as needed. The training and benchmarking are done on an NVIDIA RTX 5090. The training was done in 50 epochs for both HAM10K and KVASIR and 80 for the smoke dataset. We truncate HAM10K and KVASIR to 200 images for training and 40 images for evaluation, our fire dataset to 389 images for training and 32 for evaluation. We set the target line thickness to be 2, to ensure that the generated lines are mostly closed and dense. In addition, we use SAM 2.1~\cite{b1} as the base detector for the smoke dataset, because the set is highly noisy and not domain-specific. The learning rate we set is $10^{-4}$. The optimizer we use is AdamW\cite{b24}. We take the best EMA\cite{b23} model (which we define as having consecutively average 5 best SSIM score), with $\tau = 0.98$ at the 667th step for KVASIR and HAM10K and at the 2074th step for the Smoke dataset. We also use dropout of 0.01 for attention layers and gradient clipping of 200.0 to help prevent overfitting. Both ContourD3PM and SegRefiner were provided the exact same initial masks.

\subsection{Post-processing}
The diffusion generated contours may preserve fine geometric properties, however, the output is not guaranteed to be a closed curve and may be overly thick, due to resizing interpolation. To maximize performance, we thus introduce some post-processing techniques to ensure a closed, pixel-level aligned contour line. We apply a Gaussian blur to the image, use the Morphological Skeletonize\cite{b16} function to reduce the thick contour to a width of 1, and use the Morphological Closure function to remove the overshoots in the curve, or bridge small gaps.

 For the Smoke dataset, after applying Skeletonize\cite{b16}, we simply pick the longest, closed contour line as a binary mask output, we truncate the contour to only the areas under the smoke by taking the dot-product with a truncation mask output from the segmentation model. Please see \cref{figure3} for an illustration.

\begin{figure}[t]
\centering
  \includegraphics[width=0.45\textwidth]{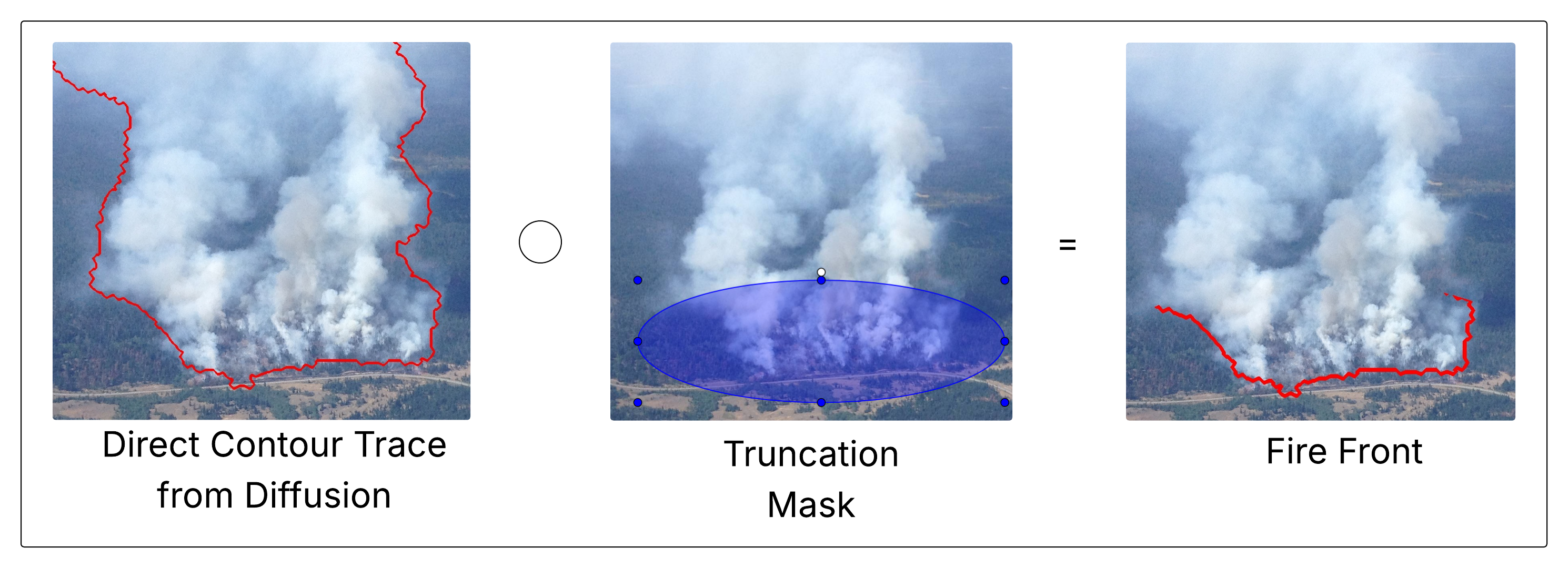}
\caption{An illustration of the boundary truncation mask. The output from our pipeline dot-producting with the truncation mask to produce the truncated contour denoting the approximate fire front. (For smoke dataset only). }
  \label{figure3}
\end{figure}

\subsection{Dataset Details}

\textbf{HAM10K}: The HAM10K dataset is a large collection of dermatoscopic images designed for research in automated skin lesion analysis. Containing 10,015 images of skin lesions from diverse populations, it includes seven common types of pigmented skin lesions, such as melanoma, basal cell carcinoma, and benign keratosis. Each image is accompanied by histopathological or expert-confirmed diagnoses, making it a valuable resource for training and evaluating machine learning models for skin cancer detection and classification.\\\\
\textbf{KVASIR}: The KVASIR data set is a comprehensive collection of images of gastrointestinal endoscopy for medical image analysis research. It contains 1,000 images from various anatomical sites within the gastrointestinal tract. Each image is verified by medical experts, providing a reliable ground truth for developing and evaluating machine learning models for automated diagnosis and classification of gastrointestinal conditions. In this dataset, we will use 200 images for training and 40 images (separated from the 200) for evaluation.\\\\
\textbf{Forest Fire (Smoke) Dataset}: We have curated our own forest fire smoke dataset in collaboration with our partners, referred to as the Smoke dataset throughout this work. The dataset includes aerial images from planes and helicopters taken by forest fire experts for rate of spread analysis. In this dataset, we will use 389 images for training and 32 images (separated from the 389) for evaluation.  

\textbf{Baseline Exclusion Note.}  We ran SegFix \cite{b27} and BPR \cite{b39}, to low-data contour refinement settings. Across multiple seeds and standard hyperparameter sweeps (learning rate, weight decay, augmentation strength), training was unstable (loss diverged / outputs collapsed to all-foreground/all-background), preventing a fair evaluation. Because these methods were originally validated on large, multi-class datasets (e.g., Cityscapes), we instead compare to SegRefiner \cite{b2} (two-stage with publicly available pretrained weights for finetuning) and MedSegDiff (diffusion segmentation baseline), both of which were reproducible in our setting. We also attempted to adapt the Bi-Directional Cascade Network (BDCN)\cite{b37} edge detector to our low-data KVASIR setting (500 train images). BDCN consistently underfit edges and overfit background texture, generalizing poorly to unseen images. We attribute this to (i) a domain gap between natural-image edge pretraining and endoscopic boundaries with low contrast/specular artifacts, and (ii) severe class imbalance from sparse edge pixels in small datasets. Because the resulting curves were unstable and far below other baselines, we exclude it from the \cref{tab:evaluation_metrics} for readability.

\begin{table*}[h]
\centering

\caption{Evaluation Metrics for Contour Refinement Across Datasets. F1: higher is better ($\uparrow$); Hausdorff
Distance, Chamfer Distance (CD), and Hausdorff Distance (HD): lower is better ($\downarrow$). (*) guide mask generator for a given dataset. Baselines written in \textit{Italics} represent the second best in a given dataset.}
\label{tab:evaluation_metrics}
\begin{tabularx}{\textwidth}{l>{\raggedright\arraybackslash}XXXXX}
\toprule
\textbf{Method} & \textbf{Dataset} & \textbf{ F1-Score}$\uparrow$ & \textbf{ Hausdorff $\downarrow$} & \textbf{ Chamfer$\downarrow$} \\
\midrule

DeepLab-v3+ \cite{b25}            & Smoke   & 0.64                 & 67.80          & 796.36\\
FCNFormer \cite{b33}             & Smoke   & {0.65}             & 77.80          & 833.46\\
SegRefiner \cite{b2} &Smoke & 0.72$\pm$0.01&57.42$\pm$ 0.49& 421.16$\pm$ 40.62\\
YOLOv11s \cite{b13}              & Smoke      & 0.72                & 75.08           & 670.00          \\
SAM2*  \cite{b1}          & Smoke      & 0.84               & {54.45}           & {307.0291}          \\
MedSegDiff   \cite{b6}          & Smoke   &{0.62$\pm$0.01}                 & 74.18$\pm$1.23           & 914.01$\pm$79.02           \\

Geodesics Active Contour\cite{b5} & Smoke   & \textbf{0.87}                &  \textbf{47.10} &  \textbf{243.07}          \\

{ContourD3PM(Ours)}     & {Smoke }& \textbf{\textit{0.85$\pm<$0.01}}          
& \textbf{\textit{49.05$\pm$0.02}}           & \textbf{\textit{273.10$\pm$0.08} }          \\

\midrule

DeepLab-v3+ \cite{b25}            & HAM   & 0.81                 & 36.24          & 509.19\\
FCNFormer   \cite{b33}           & HAM   & {0.84}                 & 31.66          & {337.98}\\

SegRefiner \cite{b2} &HAM & \textbf{0.90$\pm<$0.01}&\textbf{25.06$\pm<$ 0.01}&\textbf{155.33$\pm$ 0.30}\\
YOLOv11s*  \cite{b13}             & HAM   & {0.41}                & 27.58          & 386.59\\

SAM2   \cite{b1}          & HAM   &{0.83}                 & 49.20           & 3037.69           \\
MedSegDiff   \cite{b6}          & HAM   &{0.41$\pm$0.01}                 & 98.90$\pm$4.45           & 5874.44$\pm$4.23           \\
Geodesics Active Contour\cite{b5} & HAM   & {0.84}                &  29.86 &  {243.42}          \\

ContourD3PM(Ours)     & HAM   & \textbf{\textit{{0.86 $\pm$0.01 }  }}              &   \textbf{\textit{26.69$\pm<$0.01}  }         &   \textbf{\textit{378.16$\pm<$0.01  }}          \\

\midrule
DeepLab-v3+* \cite{b25}            & KVASIR   &  \textbf{\textit{0.83}}                   &  \textbf{\textit{24.50}}         &  {{148.31}}\\

FCNFormer \cite{b33}            & KVASIR   & 0.66                 & 30.45          & 375.32\\

SegRefiner \cite{b2} &KVASIR & 0.73$\pm$0.01&31.36$\pm$ 0.47&220.69$\pm$ 5.92\\

YOLOv11s \cite{b13}             & KVASIR   & 0.62                   & 56.31          & 1172.96             \\

SAM2  \cite{b1}           & KVASIR   & {0.71}                   & 52.23& {1147.51}           \\
MedSegDiff   \cite{b6}          & KVASIR   &{0.27$\pm$0.01}                 & 209.32$\pm$8.42           & 35002.97$\pm$7049.24           \\

Geodesics Active Contour\cite{b5} & KVASIR   & {0.82}                &  24.85 &  \textbf{\textit{144.24}}          \\

ContourD3PM(Ours)    & KVASIR   &\textbf{0.95$\pm<$0.01}  &\textbf{21.92$\pm<$0.01 }& \textbf{37.51$\pm<$0.01} \\

\bottomrule
\end{tabularx}
\end{table*}

\subsection{Qualitative Results}

Results on KVASIR, which can be seen in \cref{figure7}, demonstrate that DeepLab-v3+\cite{b25} occasionally produces sharp corners or under-detects loops. 

In our smoke dataset however, the qualitative differences between methods are not as obvious as on HAM10K, due to the fact that truncated segments of the contours are used for comparison. \\

\subsection{Quantitative Results} 

The quantitative results shows our method improves upon  base detectors in most geometric shape distance metrics. The confidence interval is at 95\% obtained via $t$ distribution after 12 evaluation runs.  SAM2.1\cite{b1} and other models are deterministic models so confidence intervals don't apply across runs. Our method improves upon all single-stage detectors, and 2 out 3 datasets for two-stage detectors in most of the shape metrics. Note that SAM2.1\cite{b1} suffers detection failures in the low data regime, causing significantly diminished performance on Hausdorff and Chamfer distance metrics, despite being competitive in many other domains.  \\
\begin{table}[h]
\centering
\caption{Ablation study on number of categories}
\resizebox{0.5\textwidth}{!}{
\label{tab:num_cat}

\begin{tabular}{lcccccc}
\toprule

\textbf{Dataset} & \textbf{State Size} & \textbf{F1$\uparrow$} & \textbf{Hausdorff$\downarrow$} & \textbf{Chamfer$\downarrow$} \\
\midrule
\multirow{4}{*}{Smoke} 
& 29 & 0.81$\pm<$0.01 & \textbf{48.66$\pm<$0.01} & 325.81$\pm<$0.01 \\
& 32 &  \textbf{0.85$\pm<$0.01}          
& {49.05$\pm$0.02}           & \textbf{273.10$\pm$0.08}  \\
& 35 & 0.69$\pm<$0.01 &  33.41$\pm<$0.01 & 408.81$\pm$0.05 \\
& 38 & 0.78$\pm<$0.01 & 72.85$\pm$0.33 & 594.52$\pm$3.82 \\
\midrule
\multirow{4}{*}{HAM10K} 
& 5  & \textbf{0.86$\pm<$0.01} & 26.85$\pm<$0.01 & 381.11$\pm<$0.01 \\
& 8  & \textbf{0.86$\pm<$0.01}  & \textbf{26.69$\pm<$0.01} & 378.55$\pm$0.01 \\
& 11 & \textbf{0.86$\pm<$0.01} & \textbf{26.69$\pm<$0.01} & \textbf{378.16$\pm<$0.01} \\
& 14 & \textbf{0.86$\pm<$0.01}  & 27.18$\pm<$0.01 & 380.25$\pm<$0.01 \\
\midrule
\multirow{4}{*}{KVASIR} 
&2  &{0.91 $\pm<$0.01}            &  {24.33$\pm$0.01}             &  {77.63$\pm$0.01}              
 \\
& 5  & 0.85$\pm<$0.01 & 25.34$\pm<$0.03 & 91.85$\pm$0.04 \\
& 8  &\textbf{0.95$\pm<$0.01}  &\textbf{21.92$\pm<$0.01 }& \textbf{37.51$\pm<$0.01} \\
& 11 & \textbf{0.92$\pm<$0.01}  & 26.93$\pm<$0.01 & 50.33$\pm$0.06 \\
\bottomrule
\end{tabular}}
\end{table}
\begin{table}[h]
\centering
\caption{Performance of our method across dataset sizes for multiple datasets. (*) used in main table.}
\label{tab:dataset_size_perf}
\resizebox{0.5\textwidth}{!}{
\begin{tabular}{llcccc}
\toprule
\textbf{Dataset} & \textbf{Samples} & F1 $\uparrow$ & Hausdorff $\downarrow$ & Chamfer $\downarrow$ \\
\midrule
\multirow{4}{*}{Smoke} 
 & 200 &  0.65$\pm$0.01 & 116.28 $\pm<$0.01 & 2651.45$\pm$9.56 \\
 & 300 & 0.79$\pm<$0.01   & 74.21$\pm<$0.01  & 607.34$\pm<$0.01 \\
 & 389* & {0.85$\pm<$0.01}          
& {49.05$\pm$0.02}           & {273.10$\pm$0.08}  \\
\midrule
\multirow{4}{*}{HAM10k} 
  & 200* &  0.86$\pm$0.01& 26.69 $\pm<$0.01 & 378.16$\pm$ 0.26 \\
 & 300 & 0.86$\pm<$0.01  & 26.72 $\pm<$0.01 & 379.28$\pm$0.03 \\
 & 400 & 0.86$\pm<$0.01  & 26.77$\pm<$0.01 & 379.07$\pm$0.04 \\
\midrule
\multirow{4}{*}{KVASIR} 
 & 200*  &{0.95$\pm<$0.01}  &{21.92$\pm<$0.01 }& {37.51$\pm<$0.01} \\
 & 300 & 0.93$\pm<$0.01 & 22.94 $\pm$0.02 & 44.48$\pm$0.03 \\
 & 400 & 0.93 $\pm<$0.01& 24.79$\pm<$0.01 &43.42$\pm$0.18 \\
\bottomrule
\end{tabular}
}
\end{table}
\begin{table}[h]
\centering
\caption{Performance of Diffusion Model under Different Denoising Steps across Three Datasets}
\label{tab:diffusion_steps_datasets}
\resizebox{0.5\textwidth}{!}{

\begin{tabular}{c|c|c|c|c}
\hline
\textbf{Dataset} & \textbf{Denoising Steps} & \textbf{F1 Score} $\uparrow$ & \textbf{Hausdorff Distance} $\downarrow$ & \textbf{Chamfer Distance} $\downarrow$  \\ 
\hline
\multirow{6}{*}{Smoke} 
& 1   & 0.84$\pm<$ 0.01  & 33.07$\pm$ 0.08 & 309.17 $\pm$ 8.96\\
& 2   & 0.85$\pm<$ 0.01  & 33.58$\pm<$ 0.01 & 287.71$\pm$ 0.15 \\
& 4   & 0.85$\pm<$ 0.01  & 52.73 $\pm<$ 0.01& 281.76 $\pm$ 0.03\\
& 8   & 0.85$\pm<$ 0.01 & 52.73$\pm<$ 0.01 & 280.72$\pm$ 0.01 \\
& 16  & 0.85 $\pm<$ 0.01 & 52.73 $\pm<$ 0.01& 280.63$\pm<$ 0.01 \\
\hline
\multirow{6}{*}{HAM10K} 
& 1   & 0.85$\pm<$ 0.01 & 26.86$\pm$ 0.02 & 377.70$\pm$ 0.18 \\
& 2   & 0.86$\pm<$ 0.01 & 26.86$\pm<$0.01 & 377.35 $\pm$0.09\\
& 4   & 0.86$\pm<$ 0.01  & 26.87$\pm<$ 0.01& 377.36 $\pm$0.03\\
& 8    & 0.86$\pm<$ 0.01 &26.87$\pm<$ 0.01 & 377.28$\pm<$ 0.01 \\
& 16  &  0.86$\pm<$ 0.01  &26.87$\pm<$ 0.01 & 377.24$\pm<$ 0.01 \\
\hline
\multirow{6}{*}{KVASIR} 
& 1   &0.94$\pm<$ 0.01 & 23.70$\pm$0.11 & 42.58$\pm$0.13 \\
& 2   & 0.94$\pm<$ 0.01 & 24.56$\pm$0.05 & 40.35$\pm$0.06 \\
& 4   & 0.94$\pm<$ 0.01  & 24.60$\pm$0.06 & 41.12$\pm$0.03 \\
& 8   & {0.94$\pm<$0.01}  &{25.18$\pm$0.02 }& {42.69$\pm$0.05}\\
&16 & {0.95$\pm<$0.01}  &{21.92$\pm<$0.01 }& {37.51$\pm<$0.01} 

\end{tabular}}
\end{table}

\begin{figure*}[h]
\centering
  \includegraphics[width=0.7\textwidth]{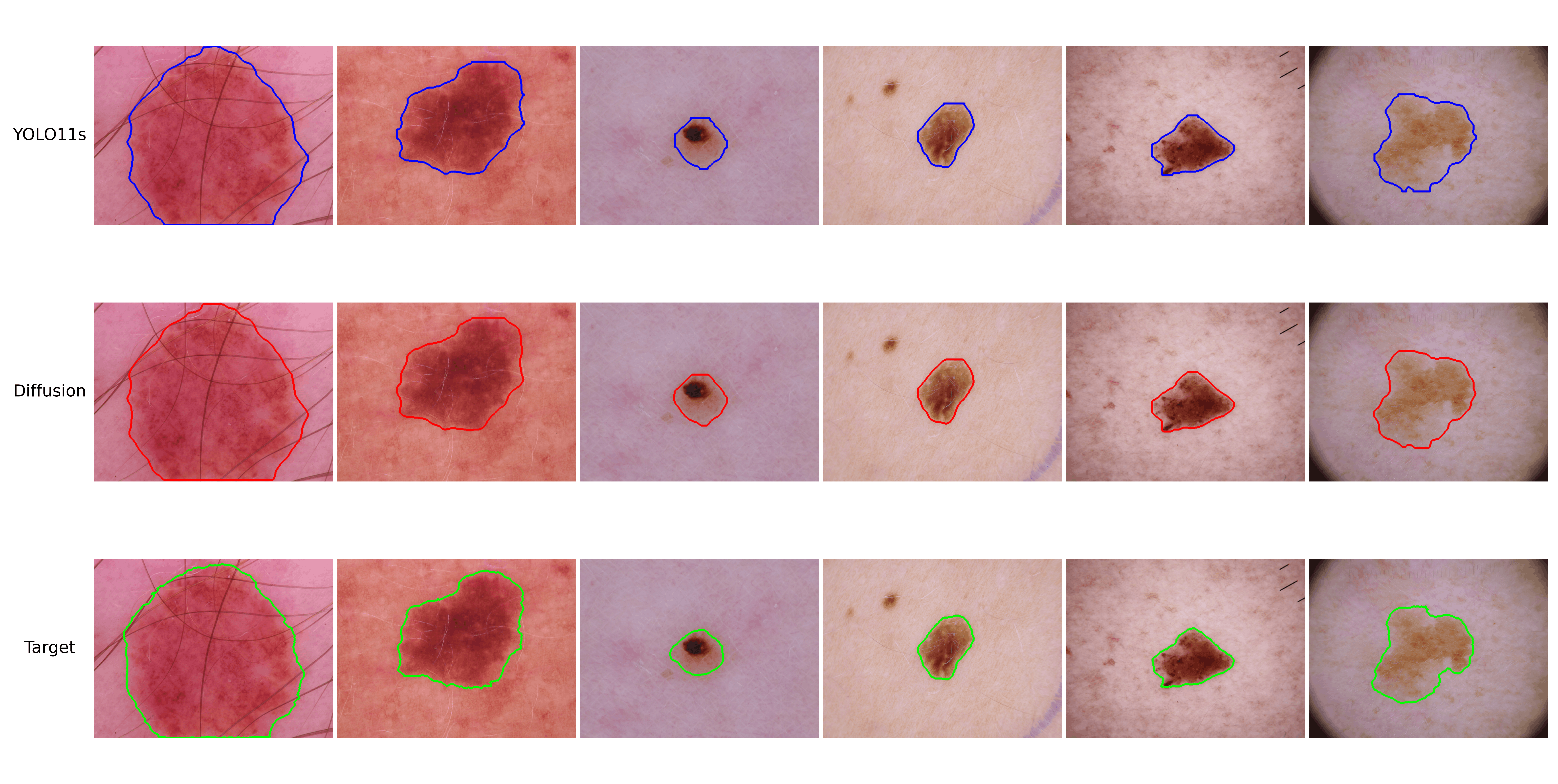}
\captionsetup{width=0.7\textwidth}
    \caption{Examples drawn from HAM10K. In this figure and following figures, the top row represents unrefined contours, the middle row represents refined and the last row represents ground truth.}
    \label{figure5}
\end{figure*}
\begin{figure*}[h]

\centering
  \includegraphics[width=0.7\textwidth]{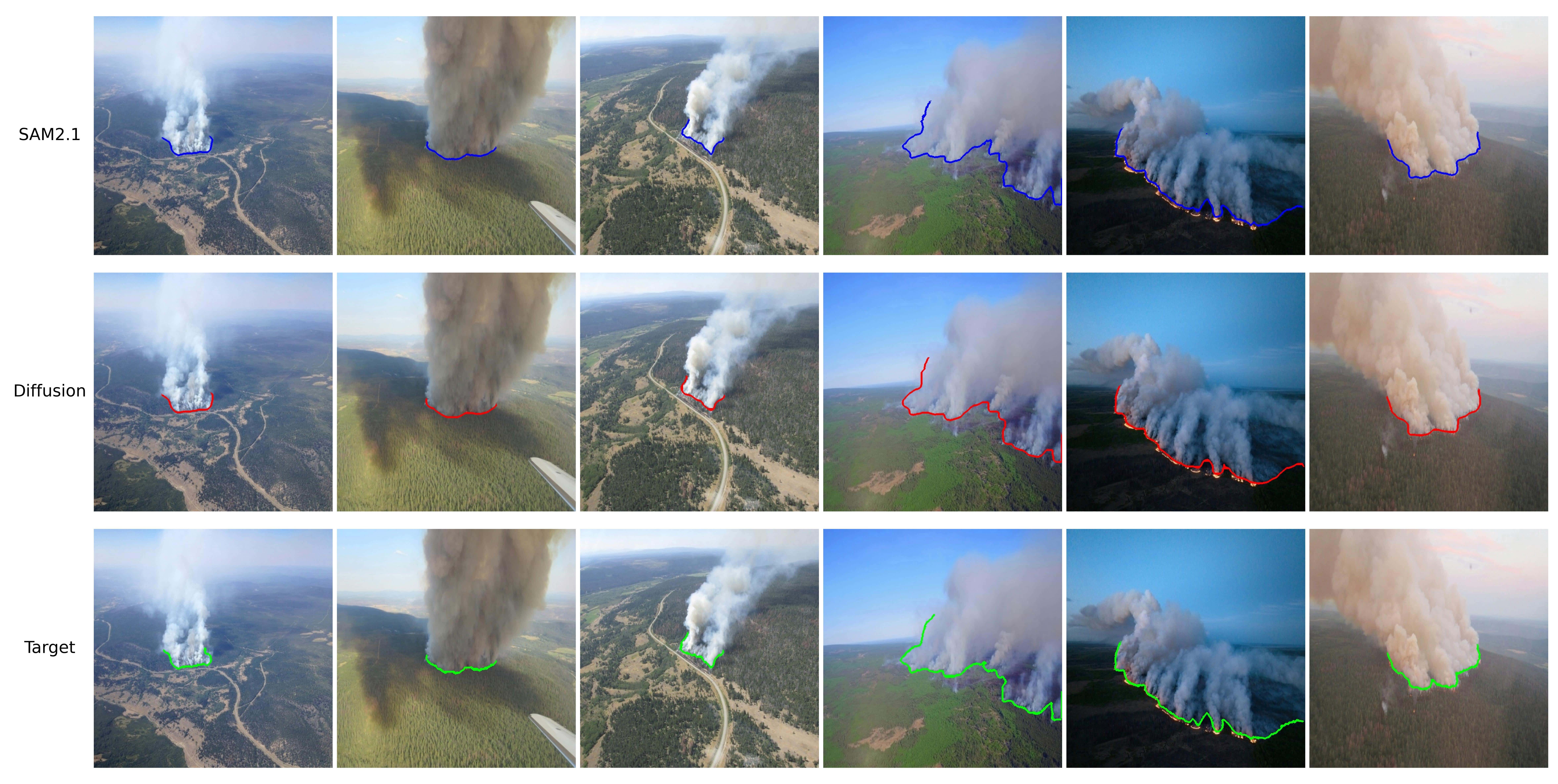}
  \captionsetup{width=0.7\textwidth}

    \caption{Examples drawn from smoke dataset. The top row rep-
resents unrefined contours, the middle row represents refined and the last row represents ground
truth.}
    \label{figure6}
\end{figure*}

\begin{figure*}[h]

\centering
  \includegraphics[width=0.7\textwidth]{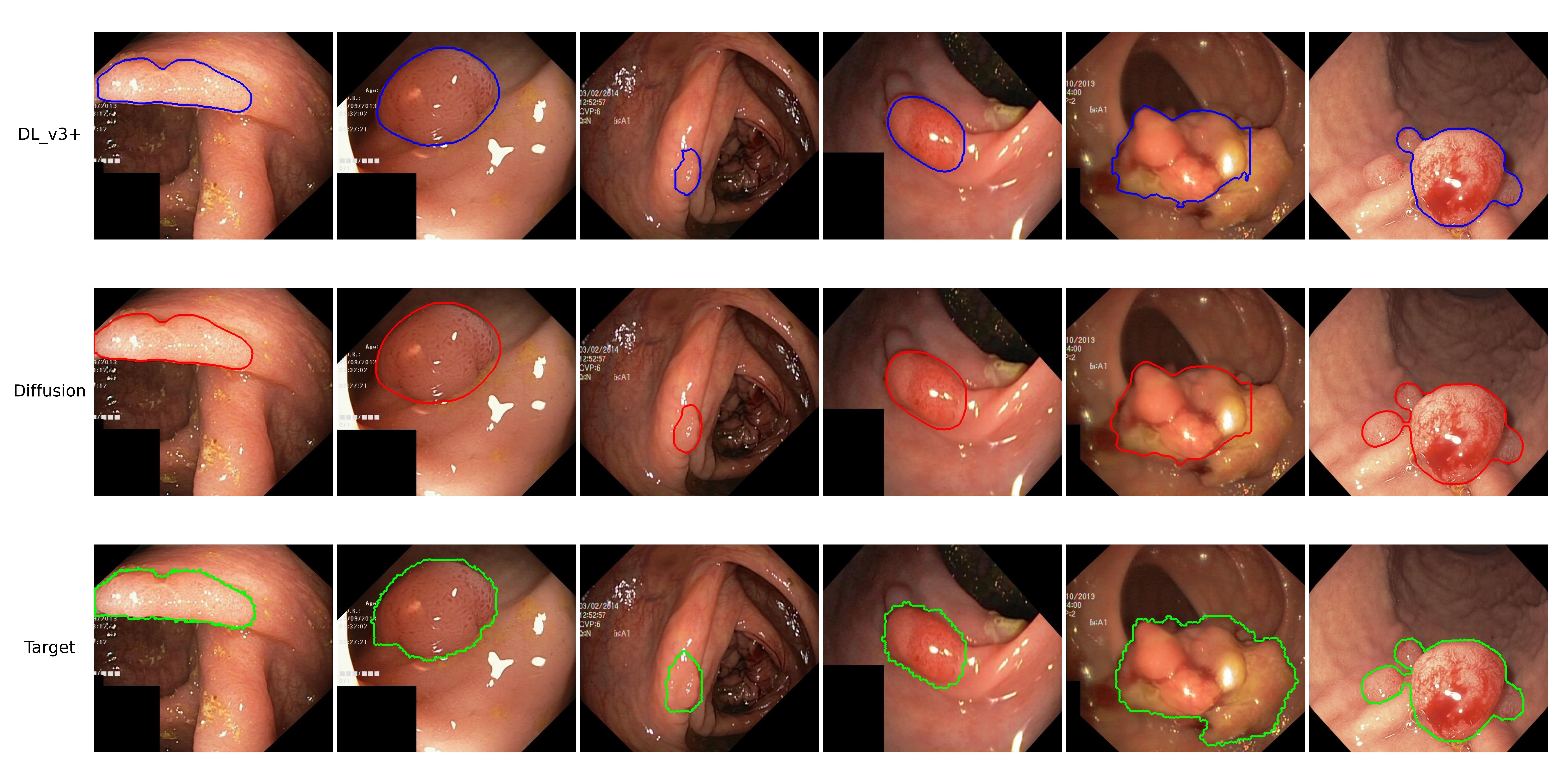}
  \captionsetup{width=0.7\textwidth}

    \caption{Examples drawn from KVASIR. The top row rep-
resents unrefined contours, the middle row represents refined and the last row represents ground
truth.}
    \label{figure7}

\end{figure*}

\begin{table}[h]
\centering
\caption{Performance of Diffusion Model using Standard vs. Simplified Reverse Process at 10 iterations}
\label{tab:diff_proc}
\resizebox{0.5\textwidth}{!}{
\begin{tabular}{c|c|c|c|c}
\hline
\textbf{Dataset} & \textbf{Method} & \textbf{F1 Score} $\uparrow$ & \textbf{Hausdorff Distance} $\downarrow$ & \textbf{Chamfer Distance} $\downarrow$  \\ 
\hline
\multirow{2}{*}{Smoke} 
& Std.  & $0.84 \pm< 0.01$ & $45.30 \pm 0.16$ & $331.70 \pm 3.21$ \\
& Simp.* & $0.85 \pm<0.01$ & $49.05 \pm 0.01$ & $273.10 \pm 0.08$ \\
\hline
\multirow{2}{*}{HAM10K~} 
& Std.  & $0.86 \pm <0.01$ & $26.79 \pm 0.03$ & $378.56 \pm 0.23$ \\
& Simp.*& $0.86 \pm <0.01$ & $26.69 \pm 0.01$ & $378.16 \pm <0.01$ \\
\hline
\multirow{2}{*}{KVASIR~} 
& Std.  & $0.94 \pm 0.01$ & $23.72 \pm 0.07$ & $40.05 \pm 0.03$ \\
& Simp.* & $0.92 \pm< 0.01$ & $24.46 \pm 0.01$ & $45.18 \pm 0.01$ \\
\hline
\end{tabular}}
\end{table}

\textbf{Ablation Studies} There are several key parameters that can potentially affect the performance of our model. We perform ablation studies on the following: size of the dataset,  number of dimensions for the state space (number of categories for confidence score), number of denoising iterations. As shown in \cref{tab:dataset_size_perf}, we experimented across 200, 300 and 400 images for dataset size. The results vary by the amount of noise in the image, where KVASIR is considered low noise and the Smoke dataset is considered to have high noise, with HAM10K somewhere in between. We found that in the smoke dataset, the model performs best at large dataset size (400). In contrast, for domain specific datasets containing less noise, the model performs slightly worse when the dataset size increases. As a stopping condition, we always stop at the 75th epoch for the smoke dataset and the 60th epoch for non-smoke dataset.  

For the confidence category studies, which can be found in  \cref{tab:num_cat}, we found that the higher the number of confidence categories used, the better the network generalizes to complex noisy images. We found that 8 categories works well in HAM10K and KVASIR, but not on the Smoke dataset, which is a dataset curated from different image sources captured by different cameras, and the background landscapes contain high variance. We resized the images to 640x640, and after observing the qualitative results during training throughout the epochs and have set the number of confidence categories significantly higher to 32. We used SAM2.1\cite{b1} as the base detector to further enhance the guide mask quality, and produced a better result than SAM2.1\cite{b1} on its own. For simplicity, we fix the training window size, as in \cite{b12}, to 352 for all datasets, and set the confidence threshold to all be $N_{cat} / 2 - 1$, where $N_{cat}$ is the number of categories.

A study of the number of denoising steps can be found in \cref{tab:diffusion_steps_datasets}. We found that increasing the number of iterations can help in narrowing the confidence intervals of the results, even though the performance sample mean may become slightly degraded. We also witnessed that increasing to above 16 denoising steps can significantly degrade performance, in both the sample mean and the confidence interval. Please note that for this ablation study, we are only testing $2^{N}-th$ iterations where $0\leq N \leq 4$ so it differs from the main table results (evaluated at 10 iterations).

\begin{table}[t]
\centering
\caption{Runtime comparison for the top-3 two staged methods(lower is better for time). SAM2.1's image size is omitted as the inference time does not depend on its input size; they're all internally resized to a fixed size. }
\label{tab:runtime_compact}
\setlength{\tabcolsep}{6pt}
\renewcommand{\arraystretch}{1.15}
\begin{tabular}{lccc}
\hline
\textbf{Method} & \textbf{Infer (s/img)}& \textbf{Num Iterations} & \textbf{Image Size}\\
\hline

SAM2.1 & 8.8720& 1 & - \\
SegRefiner & 0.5917& 6 & 256 \\

Geodesics Active Contour& 0.5230 & 150 & 352 \\
\textbf{Ours(32 Categories)} & 0.1397 & 10 & 352\\
\textbf{Ours(11 Categories)} & 0.1303 & 10 & 352\\
\textbf{Ours(8 Categories)} & 0.1328 & 10 & 352\\

\hline
\end{tabular}

\vspace{2pt}
\footnotesize
All methods are inferenced using an RTX 3060 GPU across 10 runs.\\
\underline{\hspace{0.7cm}}
\end{table}

\section{Conclusions} This work establishes discrete diffusion as a strong and practical approach for boundary-sensitive segmentation under limited supervision. Across Smoke and HAM10K, our model is consistently competitive, and on KVASIR it outperforms every baseline considered, with the largest gains appearing for occluded and translucent objects where conventional methods fail most often. We also demonstrated that our inference speed exceeded any other baselines considered, supporting applications that require real time inference and are limited in compute power. The resulting contours are not only accurate but also visually coherent, making them well-suited for applications that rely on precise delineations, including medical imaging and wildfire monitoring.

\end{document}